\documentclass[10pt,twocolumn,letterpaper]{article}

\usepackage{cvpr}                

\usepackage{cuted}
\usepackage{caption}
\usepackage{multirow}
\usepackage{booktabs}
\usepackage{threeparttable}
\usepackage{subcaption}
\usepackage{float}
\usepackage{graphicx}
\usepackage{array}
\usepackage[table]{xcolor}
\usepackage{newtxtext,newtxmath}
\usepackage{calc}
\definecolor{CitationGreen}{RGB}{116,255,116}
\definecolor{SecondBlue}{RGB}{0,112,192}
\definecolor{BestRed}{RGB}{192,0,0}
\definecolor{MissingShade}{RGB}{224,239,218}

\newcommand{\otop}{\specialrule{0.9pt}{0pt}{1pt}}
\newcommand{\omiddle}{\specialrule{0.75pt}{1pt}{1pt}}
\newcommand{\obottom}{\specialrule{0.9pt}{1pt}{0pt}}

\newcommand{\PtwoGTableWidth}{\columnwidth}

\newcommand{\PtwoGTableFont}{%
  \fontsize{7.3pt}{8.6pt}\selectfont%
  \setlength{\tabcolsep}{2pt}%
  \renewcommand{\arraystretch}{1.05}%
}

\definecolor{cvprblue}{rgb}{0.21,0.49,0.74}
\usepackage[
    pagebackref,
    breaklinks,
    colorlinks,
    allcolors=cvprblue
]{hyperref}

\def\confName{CVPR}
\def\confYear{2027}

\title{Poisson2Gaussian: Self-supervised Noise Gaussianization 
\\to Enhance Image Denoising}

\author{
Xirou Zhou$^{1}$ \quad
Zijing Xu$^{1}$ \quad
Yibo Qu$^{2}$ \quad
Qi Zhang$^{2}$ \quad
Xiaowan Hu$^{3*}$ \quad
Xinyang Li$^{1*}$ \\
{\small $^{1}$College of AI, Tsinghua University, Beijing, China} \\
{\small $^{2}$Department of Automation, Tsinghua University, Beijing, China} \\
{\small $^{3}$School
of Electronic and Information Engineering, Beihang University, Beijing, China} \\
{\small $^{*}$Correspondence: huxiaowan@buaa.edu.cn; xinyangli@tsinghua.edu.cn}
}

\begin{document}
\maketitle

\begin{abstract}
The quantum nature of light gives rise to Poisson shot noise, a fundamental source of noise in photography, microscopy, and astronomy. Our controlled experiments reveal that denoising performance remains sensitive to noise distributions even under matched noise strength and in-domain self-supervised training. Motivated by this finding, we propose Poisson2Gaussian (P2G), a learnable transformation that uses an invertible flow to reshape the joint distribution of estimated noise toward i.i.d.\ Gaussian noise. However, Gaussian likelihood alone cannot determine the denoising utility of a transformation, as identical likelihoods can yield different restoration outcomes. We therefore introduce a self-supervised framework that combines likelihood-based Gaussianization with denoising-oriented matching and alternating optimization. For independent, conditionally unbiased noisy pairs, the expected matching loss differs from supervised MSE by a model-independent constant, while approximate pairs enable training from a single noisy image. P2G improves denoising on FMDD, W2S, and SIDD RAW and sRGB, outperforming paired-noisy self-supervised baselines on FMDD by up to 0.80\,dB. Consistent gains across UNet and NAFNet further demonstrate its effectiveness with different denoiser architectures. Code will be publicly available.
\end{abstract}

\begin{figure}[t]
    \centering
    \includegraphics[width=\columnwidth]{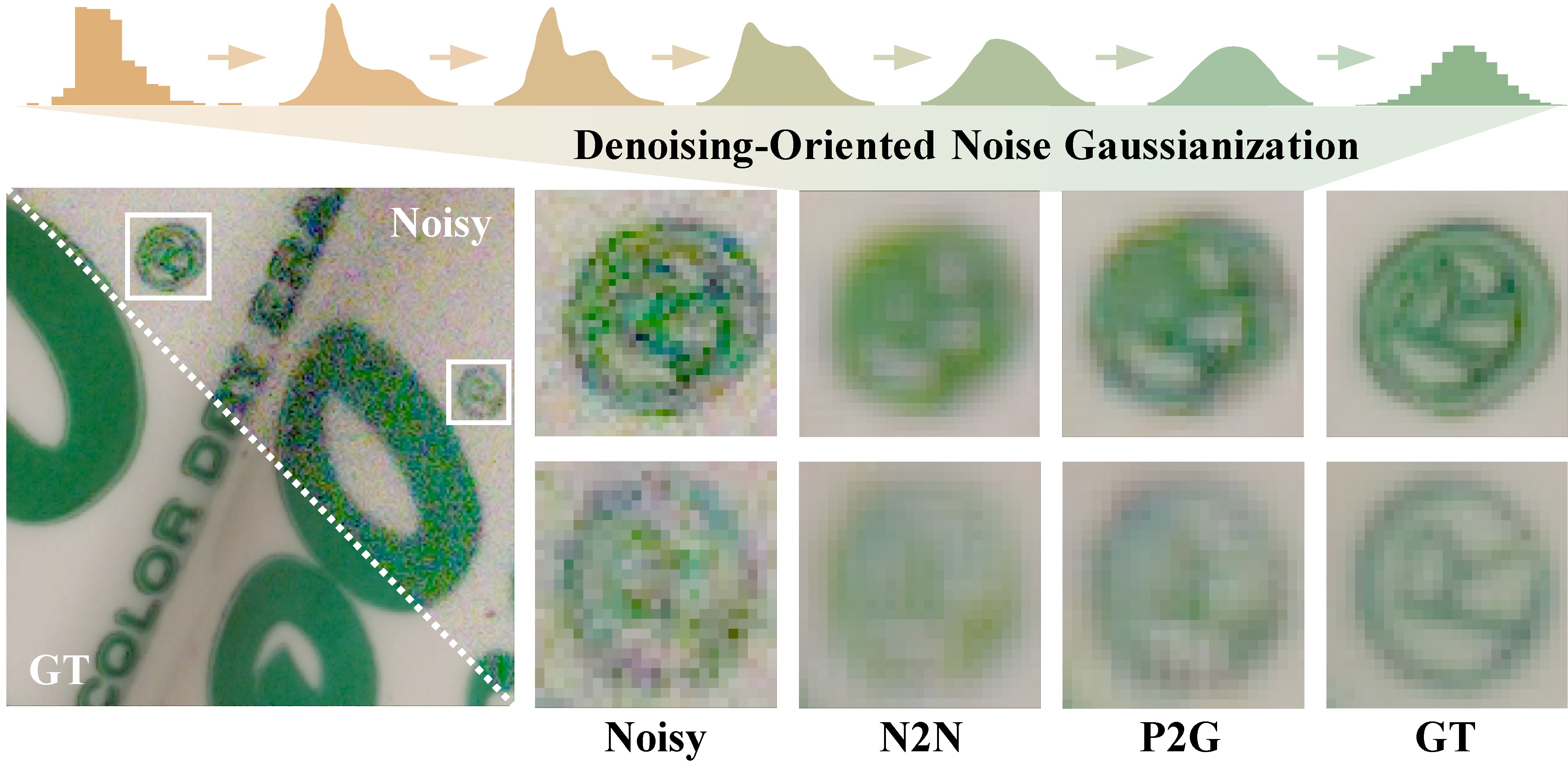}
    \caption{\textbf{P2G improves denoising through self-supervised noise Gaussianization.}
    P2G learns an invertible transformation that maps estimated noise toward i.i.d.\ Gaussian noise, motivated by signal-dependent variance and non-Gaussian statistics. A self-supervised framework couples P2G with a trainable denoiser to align Gaussianization with downstream denoising performance.}
    \label{fig1}
\end{figure}

\section{Introduction}
\label{sec:intro}

\begin{figure*}[t]
  \centering
  \includegraphics[width=1\textwidth]{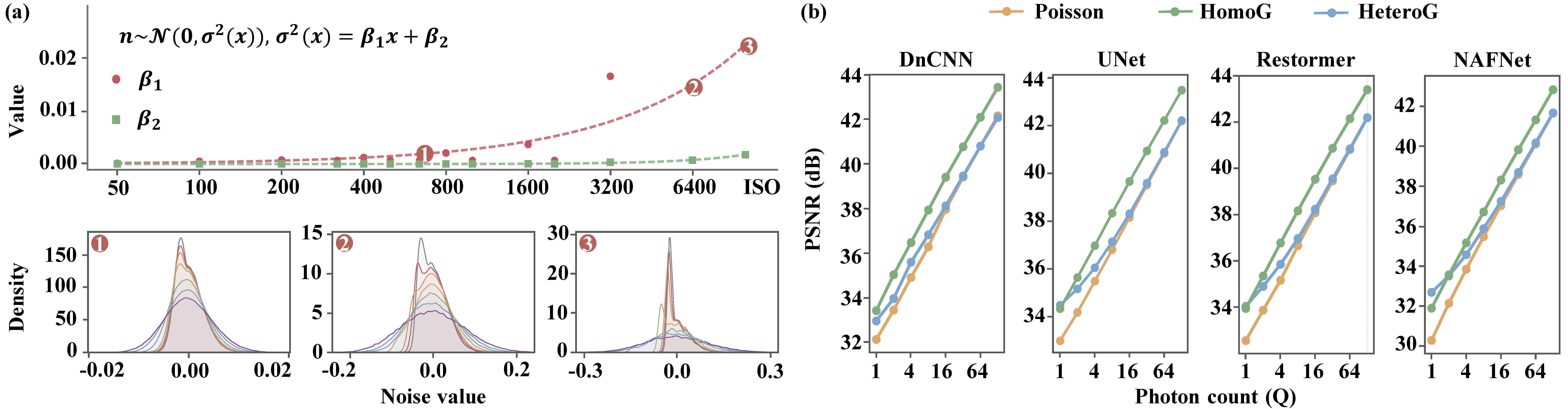}
  \caption{\textbf{Rationale for noise Gaussianization.}
    (a) Noise statistics in real camera data from SIDD. The signal-dependent variance coefficient $\beta_1$ generally increases with ISO, while representative residual histograms exhibit non-Gaussian characteristics, including asymmetry. 
    (b) Denoising PSNR of four architectures trained under the same N2N~\cite{lehtinen2018noise2noise} protocol on synthetic FMDD at photon counts $Q$. Poisson and heteroscedastic Gaussian (HeteroG) noise have matched conditional means and per-pixel variances, while homoscedastic Gaussian (HomoG) noise is matched in global noise strength. Input PSNR and training conditions are matched across noise types. Performance differences persist despite in-domain training, including between Poisson and HeteroG noise at low photon counts.}
  \label{fig2}
\end{figure*}

Image denoising is fundamental to computer vision, enabling reliable visualization and downstream analysis. Advances in imaging hardware have reduced electronic noise, increasing the relative importance of photon shot noise beyond traditionally photon-limited applications such as low-light photography, microscopy, and astronomy. Arising from the quantum nature of photon detection, Poisson shot noise exhibits signal-dependent variance and non-Gaussian statistics, particularly at low photon counts~\cite{casacio2021quantum,li2023real,li2021reinforcing,samantaray2017realization}. These properties are also reflected in the SIDD dataset~\cite{abdelhamed2018sidd}, where signal-dependent noise coefficients generally increase with ISO and representative noise residuals deviate from Gaussianity (\cref{fig2}a). Despite advances in learning-based denoising~\cite{li2023spatial,elad2023imagedenoising}, complex noise statistics remain a challenge for image restoration.

Existing approaches address complex noise through noise transformation or noise-aware learning. Classical variance-stabilizing transformations (VSTs) facilitate Gaussian denoising~\cite{foi2008practical,AzzariFoi2016VSTPoisson}, while FBI-Denoiser~\cite{byun2021fbi} incorporates this principle into learning-based restoration. LAN~\cite{kim2024lan} and Noise2VST~\cite{Herbreteau2025Noise2VST} learn transformations for pretrained Gaussian denoisers, while NTN~\cite{Ha2026NTN} further introduces explicit distributional constraints and supervised restoration objectives. In contrast, self-supervised denoising enables in-domain training without clean targets or a pretrained Gaussian denoiser. For independent, conditionally unbiased noisy pairs, N2N~\cite{lehtinen2018noise2noise} shows that noisy-target MSE differs from supervised MSE in expectation only by a model-independent constant, without requiring an explicit noise model. Blind2Sound~\cite{Liu2025Blind2Sound} uses Poisson--Gaussian noise estimates and adaptive objectives to improve self-supervised denoising. Yet the ability to train without an explicit noise model does not make denoising insensitive to the input noise distribution. This raises a question: even with in-domain self-supervised training, can noise transformation further improve restoration, and which statistics matter?

Our controlled experiments investigate this dependence across Poisson, heteroscedastic Gaussian, and homoscedastic Gaussian noise (\cref{fig2}b). Under matched input PSNR and training conditions, homoscedastic Gaussian noise generally yields higher denoising PSNR, particularly at moderate and high photon counts. At low photon counts, Poisson and heteroscedastic Gaussian noise yield different restoration performance despite matched conditional means and per-pixel variances (Supp. Sec.~1, Fig.~S1, and Tab.~S1).
These findings show that noise distributions affect denoising performance even under in-domain training, motivating deliberate noise transformation rather than relying on denoiser adaptation alone.

Accordingly, we introduce Poisson2Gaussian (P2G), which uses an invertible flow to reshape the joint distribution of estimated noise toward i.i.d.\ Gaussian noise without prescribing a Poisson--Gaussian noise model. Gaussian likelihood provides an exactly computable objective for the joint distribution of transformed residuals, but cannot identify the transformation most beneficial for denoising. Post-flow orthogonal transformations preserve likelihood yet can produce different restoration outcomes with a fixed denoiser (\cref{fig5}). This ambiguity motivates guiding Gaussianization with downstream denoising feedback.

To obtain this feedback when clean targets are unavailable, as often occurs in photon-limited imaging, we couple P2G with a trainable denoiser in a self-supervised framework. Gaussian NLL constrains the joint distribution of transformed residuals, while self-supervised matching provides denoising-oriented feedback. For independent, conditionally unbiased noisy pairs, the expected matching loss differs from supervised MSE of the composite prediction by a model-independent constant~\cite{lehtinen2018noise2noise}. Approximate noisy pairs enable training from a single image~\cite{huang2021neighbor2neighbor}. Alternating updates allow P2G to follow denoising feedback and the denoiser to adapt to transformed inputs.

Extensive experiments on synthetic data and real-world benchmarks, including FMDD, W2S, and SIDD (RAW and sRGB), demonstrate the effectiveness of P2G under both paired- and single-observation settings. On FMDD, P2G achieves up to 0.80\,dB higher PSNR than existing paired-noisy self-supervised methods.

Our main contributions are:
\begin{itemize}

    \item We reveal that noise distributions affect in-domain self-supervised denoising even under matched noise strength or matched conditional means and variances.

    \item We introduce P2G to learn joint Gaussianization of estimated noise residuals without a prescribed Poisson--Gaussian noise model.

    \item We make Gaussianization denoising-oriented through self-supervised matching and coupled adaptation of the transformation and denoiser.

    \item We demonstrate consistent improvements across synthetic and real-world datasets, observation settings, and denoiser architectures.
\end{itemize}

\section{Related Work}\label{related}
\subsection{Noise Models}\label{noisemodels}
Classical image denoising often assumes additive white Gaussian noise (AWGN)~\cite{buades2005non,aharon2006ksvd,dabov2007image_bm3d,zoran2011epll}. Photon-limited observations are commonly described by mixed Poisson-Gaussian noise (MPGN), combining signal-dependent shot noise with Gaussian readout noise~\cite{foi2008practical,luisier2010image_purelet}. Real camera noise can also exhibit spatial and channel dependencies introduced by sensor characteristics and processing pipelines~\cite{Abdelhamed2019NoiseFlow,Wei2020ELD,Kousha2022sRGBFlow,LLD}. These properties motivate examining denoising sensitivity to noise distributions beyond overall noise strength.

\subsection{Noise Adaptation}
Variance-stabilizing transformations (VSTs) map Poisson and Poisson-Gaussian observations toward approximately homoscedastic Gaussian noise~\cite{Anscombe1948Biometrika,foi2008practical}; accurate inversion is important after nonlinear transformation~\cite{Makitalo2011OptInvAnscombe,Makitalo2011ClosedForm,makitalo2012optimal}. FBI-Denoiser~\cite{byun2021fbi}, Neighbor2Global~\cite{Chen2024Neighbor2Global}, and SPGM-Net~\cite{Wang2026SPGM} bring GAT-based noise processing into learning-based denoising, including self-supervised settings. Related work improves VST calibration and blind RAW noise estimation~\cite{YOND2026,Wang2026RPGVST}. Unlike these model-based approaches, P2G learns to transform the joint distribution of estimated noise residuals without assuming a fixed Poisson--Gaussian noise model.

Learned input adaptation offers a different route. LAN~\cite{kim2024lan} learns pixel-wise offsets for a fixed pretrained denoiser; Noise2VST~\cite{Herbreteau2025Noise2VST} learns a transform guided by a pretrained Gaussian denoiser without updating its parameters. NTN~\cite{Ha2026NTN} combines distributional constraints with a supervised restoration objective for noise translation. Fu et al.~\cite{fu2025transforming} use histogram-based transformations and iterative refinement to extend a pretrained Gaussian denoiser to new noise conditions, while spatial rearrangement has also been used to adapt AWGN denoisers to real noise~\cite{zhou2020awgn}. These works establish that noise transformations can benefit restoration. P2G instead learns to transform the joint residual distribution alongside an in-domain self-supervised denoiser, using Gaussian likelihood and denoising feedback.

\subsection{Noise Flow}
Normalizing flows provide invertible density models with tractable likelihoods~\cite{dinh2014nice,Dinh2017RealNVP,glow}. Noise Flow~\cite{Abdelhamed2019NoiseFlow} and subsequent camera-noise models use conditional flows to capture complex noise statistics~\cite{noisee2noiseflow,Kousha2022sRGBFlow,naflow}. Notably, Noise2NoiseFlow~\cite{noisee2noiseflow} jointly trains a conditional noise model and a denoiser using noisy pairs, demonstrating that flow modeling and self-supervised denoiser learning can be combined. Whereas Noise2NoiseFlow models the conditional noise distribution, P2G transforms estimated residuals to construct the denoiser input, with self-supervised restoration feedback guiding that transformation.

\subsection{Self-Supervised Denoising}
Self-supervised denoising learns without paired clean targets. Noise2Noise~\cite{lehtinen2018noise2noise} exploits independent noisy observations, while single-image methods construct learning signals through subsampling, masking, perturbations, or consistency~\cite{krull2019noise2void,huang2021neighbor2neighbor,zs-n2n,wang2022blind2unblind,Pan_2023_sdap,Chihaoui_2024_CVPR_mash,Li2025Positive2Negative}. Other methods address spatial correlations and noise variability through specialized network designs or losses~\cite{lee2022apbsn,Liu2025Blind2Sound,Shan2026NSP,Park2026TMBSN}. In particular, Blind2Sound~\cite{Liu2025Blind2Sound} uses noise-level-aware adaptive re-visible training and a Gaussian regularizer to improve Poisson-Gaussian denoising. Unlike methods that primarily modify the denoiser or its training objective, P2G explicitly learns a noise transformation guided by self-supervised denoising feedback.

\section{Method}
\label{sec:method}
\subsection{Noise Model and Residual Transformation}\label{motivation}

\begin{figure*}[t]
  \centering
  \includegraphics[width=1\textwidth]{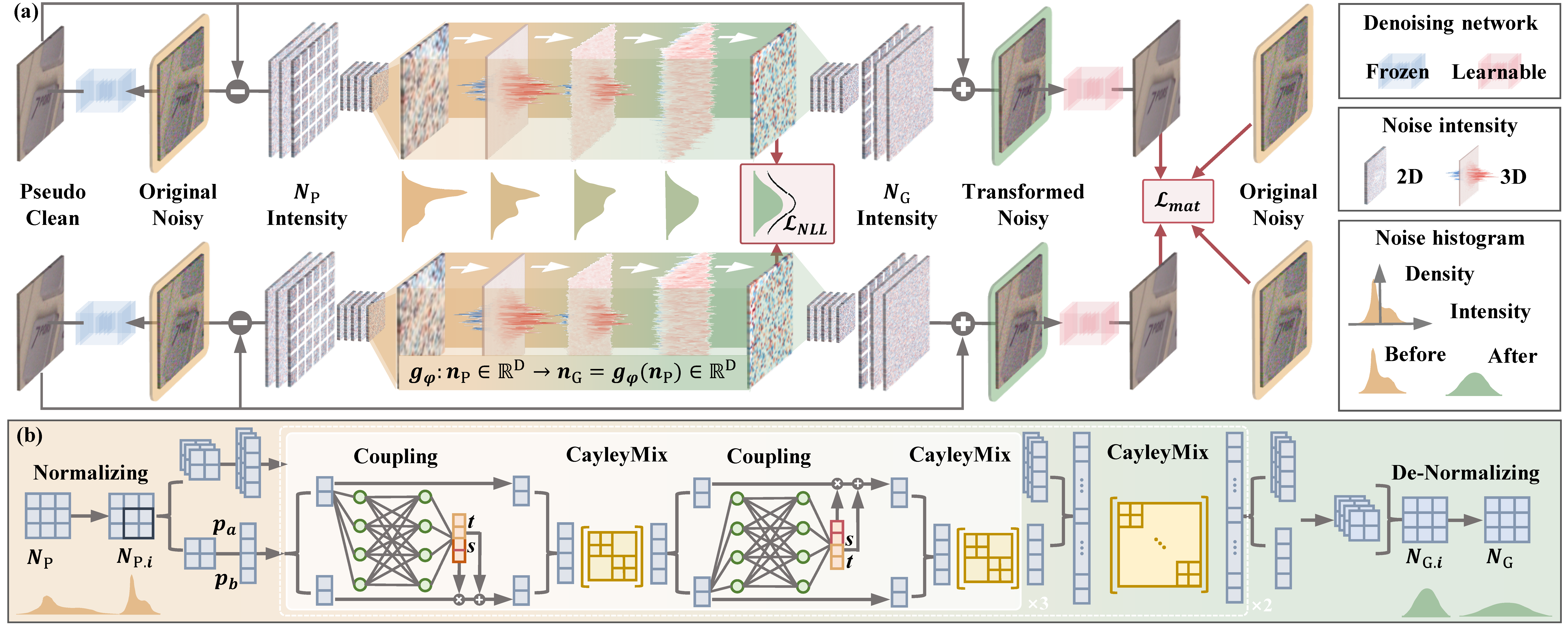}
  \caption{\textbf{Denoising-oriented Gaussianization framework.}
  (a) A denoiser estimates a pseudo-clean image, from which P2G transforms the estimated residual before reconstruction of the denoiser input. The transformation and denoiser are optimized alternately: P2G uses Gaussian NLL and self-supervised matching, while denoiser updates use matching alone.
  (b) The invertible flow comprises alternating affine coupling and CayleyMix layers.}
  \label{fig3}
\end{figure*}

\noindent\textbf{Noise model and controlled evidence.}
Photon-limited imaging is commonly described by mixed Poisson--Gaussian noise (MPGN), combining shot noise and Gaussian readout noise:
\begin{equation*}
\begin{aligned}
Y\mid X &\sim \alpha\,\mathcal{P}\!\left(\frac{X}{\alpha}\right)+\mathcal{N}(0,\beta),\\
\mathbb E[Y\mid X]&=X,\qquad \mathrm{Var}[Y\mid X]=\alpha X+\beta,
\end{aligned}
\end{equation*}
where $\mathcal{P}$ and $\mathcal{N}$ denote the Poisson and Gaussian distributions, respectively; $X$ is the clean signal, $\alpha>0$ the sensor gain, and $\beta$ the Gaussian noise variance. This model exhibits signal-dependent variance and non-Gaussian statistics at low photon counts. The SIDD coefficients and representative residual histograms in \cref{fig2}a illustrate the relevance of these properties to real images.

The controlled comparisons in \cref{fig2}b use in-domain Noise2Noise (N2N) training~\cite{lehtinen2018noise2noise} and matched noisy-input PSNR (Supp. Sec.~1 and Tab.~S1). Even matching conditional moments between Poisson and HeteroG noise does not equalize denoising performance. This motivates learning the residual distribution presented to the denoiser, beyond variance stabilization alone.

Given an observation $Y$, let $\hat X=f_\theta(Y)$ be the current denoiser estimate and $\sigma=\mathrm{std}(Y-\hat X)>0$. P2G transforms the normalized estimated residual $N_{\mathrm P}=(Y-\hat X)/\sigma$ into $N_{\mathrm G}=g_\phi(N_{\mathrm P})$, using an invertible flow $g_\phi$.

\subsection{Poisson2Gaussian}\label{poisson2gaussian}

\noindent\textbf{Transformation and recoverability.}
The transformed denoiser input is $\hat Y=\hat X+\sigma N_{\mathrm G}$. The full augmented representation remains recoverable, even though the denoiser receives only $\hat Y$:
\begin{equation*}
T(Y)=(\hat X,N_{\mathrm G},\sigma).
\end{equation*}
For fixed model parameters and an invertible $g_\phi$, the original observation is recoverable from this representation:
\begin{equation*}
Y=\hat X+\sigma g_\phi^{-1}(N_{\mathrm G}).
\end{equation*}
Thus $I(X;T(Y))=I(X;Y)$ under these conditions. This does not establish information preservation for $\hat Y$ alone or guarantee improved restoration; the precise conditions and proof are given in Supp. Sec.~2.

\noindent\textbf{Invertible flow.}
P2G interleaves affine coupling layers~\cite{Dinh2017RealNVP}, which provide nonlinear invertible mappings with tractable Jacobians, and CayleyMix layers, which mix residual dimensions orthogonally. For an orthogonal matrix $U$, $U^{-1}=U^\top$ and $\log|\det U|=0$, so this mixing adds no Jacobian log-determinant contribution (Supp. Sec.~3).

\noindent\textbf{Gaussian likelihood.}
Normalized residual patches are flattened into $\mathbf n_{\mathrm P}\in\mathbb R^D$ and transformed as $\mathbf n_{\mathrm G}=g_\phi(\mathbf n_{\mathrm P})$. By change of variables, the Gaussian negative log-likelihood (NLL) is
\begin{equation*}
\mathcal L_{\mathrm{NLL}}=
\frac12\|\mathbf n_{\mathrm G}\|_2^2
-\log\left|\det J_{g_\phi}(\mathbf n_{\mathrm P})\right|
+\frac D2\log(2\pi).
\end{equation*}
For a fixed input residual distribution, minimizing expected NLL is equivalent, up to a constant, to minimizing the KL divergence of the transformed distribution from $\mathcal{N}(0,I)$. Thus, the objective targets the joint residual density rather than only its variance.

Crucially, Gaussian likelihood cannot uniquely select the transformation most useful for denoising. For any orthogonal matrix $R$ applied after the flow,
\begin{equation*}
\mathcal L_{\mathrm{NLL}}(R\circ g_\phi;\mathbf n_{\mathrm P})
=\mathcal L_{\mathrm{NLL}}(g_\phi;\mathbf n_{\mathrm P}).
\end{equation*}
Yet these transformations can produce different denoiser inputs and restoration outcomes, as verified in \cref{fig5}. Likelihood-based Gaussianization therefore requires additional guidance from the denoising task.

\subsection{Self-Supervised Synergistic Framework}\label{4s}

We couple P2G with a trainable denoiser so that the transformation is guided by both Gaussian likelihood and denoising performance. Two training settings share this framework: P2G-2o uses independent noisy observations of the same signal, whereas P2G-1o forms approximate noisy pairs from a single image following Neighbor2Neighbor (N2Nbr)~\cite{huang2021neighbor2neighbor}.

\noindent\textbf{Denoising-oriented matching.}
For each noisy observation $Y_i$, the composite prediction is formed from
\begin{equation*}
\begin{gathered}
\hat X_i=f_\theta(Y_i),\qquad
\sigma_i=\mathrm{std}(Y_i-\hat X_i),\\
N_{\mathrm P,i}=\frac{Y_i-\hat X_i}{\sigma_i},\quad
N_{\mathrm G,i}=g_\phi(N_{\mathrm P,i}),\quad
\hat Y_i=\hat X_i+\sigma_i N_{\mathrm G,i}.
\end{gathered}
\end{equation*}
We optimize downstream denoising through the symmetric matching loss
\begin{equation*}
\mathcal L_{\mathrm{mat}}
=\|f_\theta(\hat Y_1)-Y_2\|_2^2
+\|f_\theta(\hat Y_2)-Y_1\|_2^2.
\end{equation*}
In the two-observation setting, suppose $\mathbb E[Y_i\mid X]=X$ and $Y_1\perp\!\!\!\perp Y_2\mid X$. Because $\hat Y_1$ depends only on $Y_1$ for fixed parameters, the N2N identity~\cite{lehtinen2018noise2noise} gives
\begin{equation*}
\begin{aligned}
\mathbb E\|f_\theta(\hat Y_1)-Y_2\|_2^2
={}&\mathbb E\|f_\theta(\hat Y_1)-X\|_2^2+\mathbb E\|Y_2-X\|_2^2.
\end{aligned}
\end{equation*}
The second term is independent of $\theta$ and $\phi$, and the reverse term obeys the same relation. Thus, expected matching loss differs from supervised MSE of the composite prediction by a model-independent constant. This exact equivalence applies to independent, conditionally unbiased noisy pairs; the single-observation N2Nbr pairs are an approximation. NLL supplies a distributional objective, while matching makes Gaussianization responsive to denoising quality.

\begin{figure*}[t]
  \centering
  \includegraphics[width=1\textwidth]{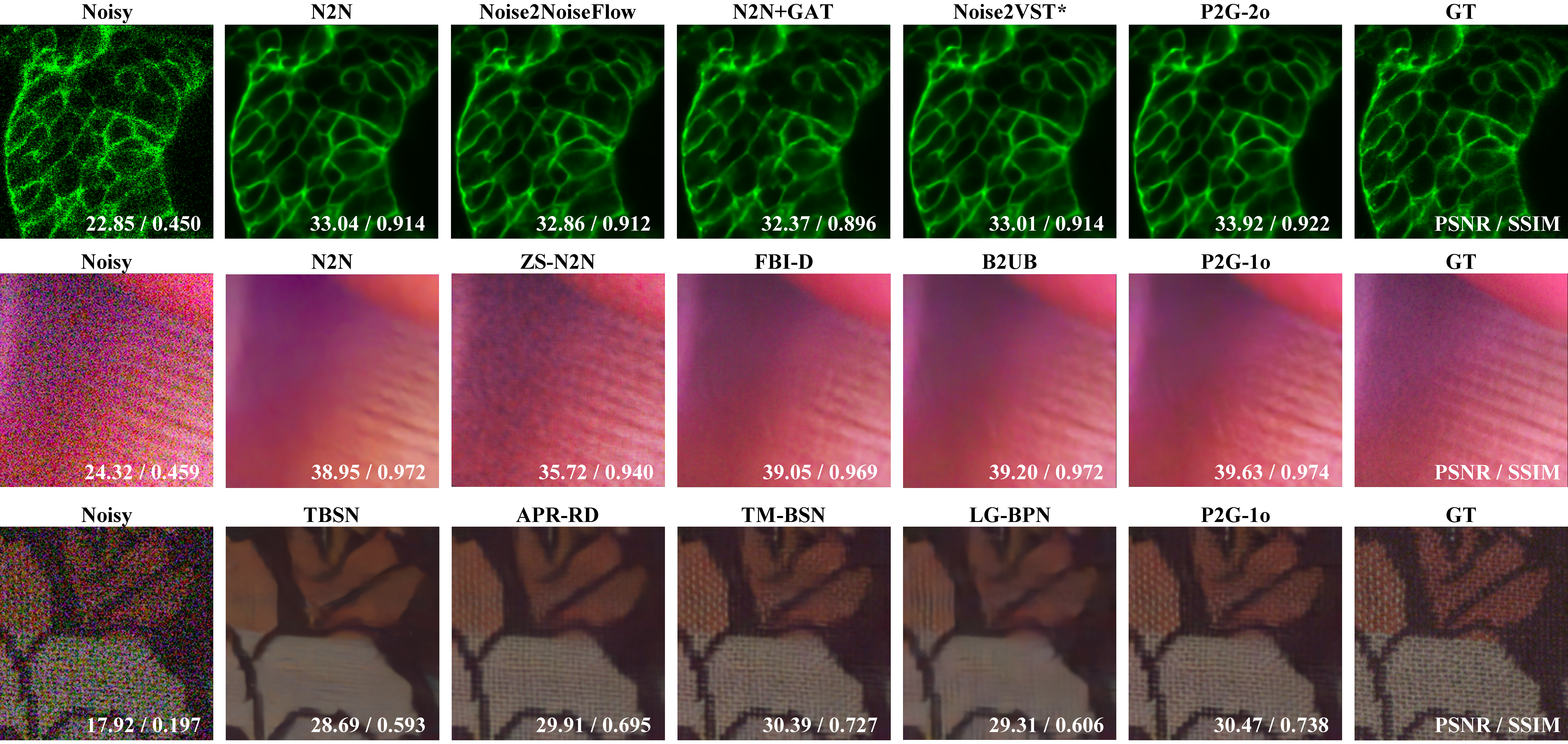}
  \caption{\textbf{Visual denoising comparisons on FMDD and SIDD.} Rows: FMDD, SIDD RAW, and SIDD sRGB. PSNR/SSIM are measured on full images; RAW is rendered in sRGB for display.}
  \label{fig6}
\end{figure*}

A separate experiment on synthetic RGB images (Supp. Sec.~4, Fig.~S2 and Tab.~S2) shows that denoising PSNR varies by 1.53--2.07\,dB across input noise distributions, but by at most 0.21\,dB across target noise distributions for a fixed input, over the tested photon counts. This asymmetry motivates transforming the denoiser input while retaining the original noisy observation as the matching target.

\noindent\textbf{Alternating optimization.}
The two components are optimized in alternating stages. With $f_\theta$ fixed, P2G is updated using Gaussian NLL and matching; with $g_\phi$ fixed, the denoiser is updated using matching alone. During denoiser updates, gradients pass through the recomputed $\hat X_i$, $\sigma_i$, and $\hat Y_i$ to $\theta$. This arrangement allows the transformation to follow the current residual estimates and denoising performance, while the denoiser adapts to the transformed inputs.

\section{Experiments}
We evaluate noise Gaussianization, the non-uniqueness of likelihood-based transformations, real-data denoising, and component contributions through controlled experiments, benchmarks, and ablations.

\subsection{Settings}\label{settings}
\noindent\textbf{Datasets and protocols.}
Synthetic tests use FMDD ground truths from 12 samples, with FOVs 3--20 for training and FOVs 1--2 for testing at $Q\in\{1,4,16,64\}$. Real-data benchmarks use FMDD~\cite{zhang2018fmdd}, W2S~\cite{zhou2020w2s}, and SIDD~\cite{abdelhamed2018sidd}. FMDD and W2S comparisons follow Noise2VST~\cite{Herbreteau2025Noise2VST}; the FMDD real-data test uses FOV 19, separate from the synthetic test split. SIDD RAW Benchmark scores are obtained from the official server; RAW and sRGB Validation use their respective official protocols. Backbone experiments use separate local splits (\cref{tab:backbone}).

\begin{figure*}
  \centering
  \includegraphics[width=1\textwidth]{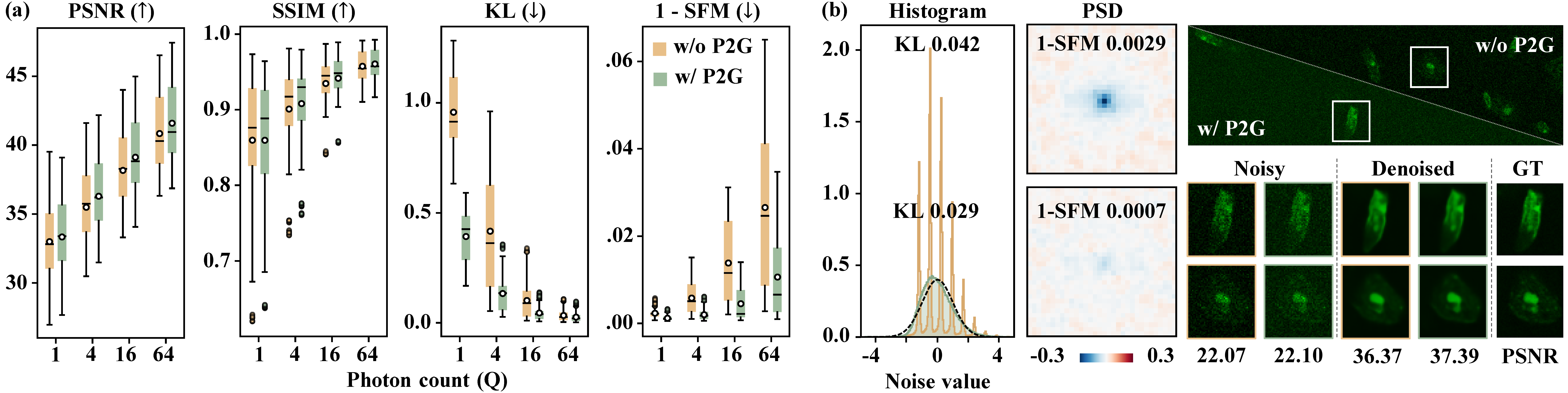}
  \caption{\textbf{P2G improves residual Gaussianization and denoising performance across photon counts.}
 (a) PSNR, SSIM, marginal KL divergence, and $1-\mathrm{SFM}$ across photon counts. (b) Residual histograms, power spectral density (PSD), and representative images at $Q=16$. Lower KL and $1-\mathrm{SFM}$ indicate closer agreement with Gaussian and white-noise statistics. The baseline is pretrained N2N; results labeled P2G use the fully adapted flow--denoiser framework.}
  \label{fig4}
\end{figure*}

\begin{figure}[t]
  \centering
  \includegraphics[width=1\linewidth]{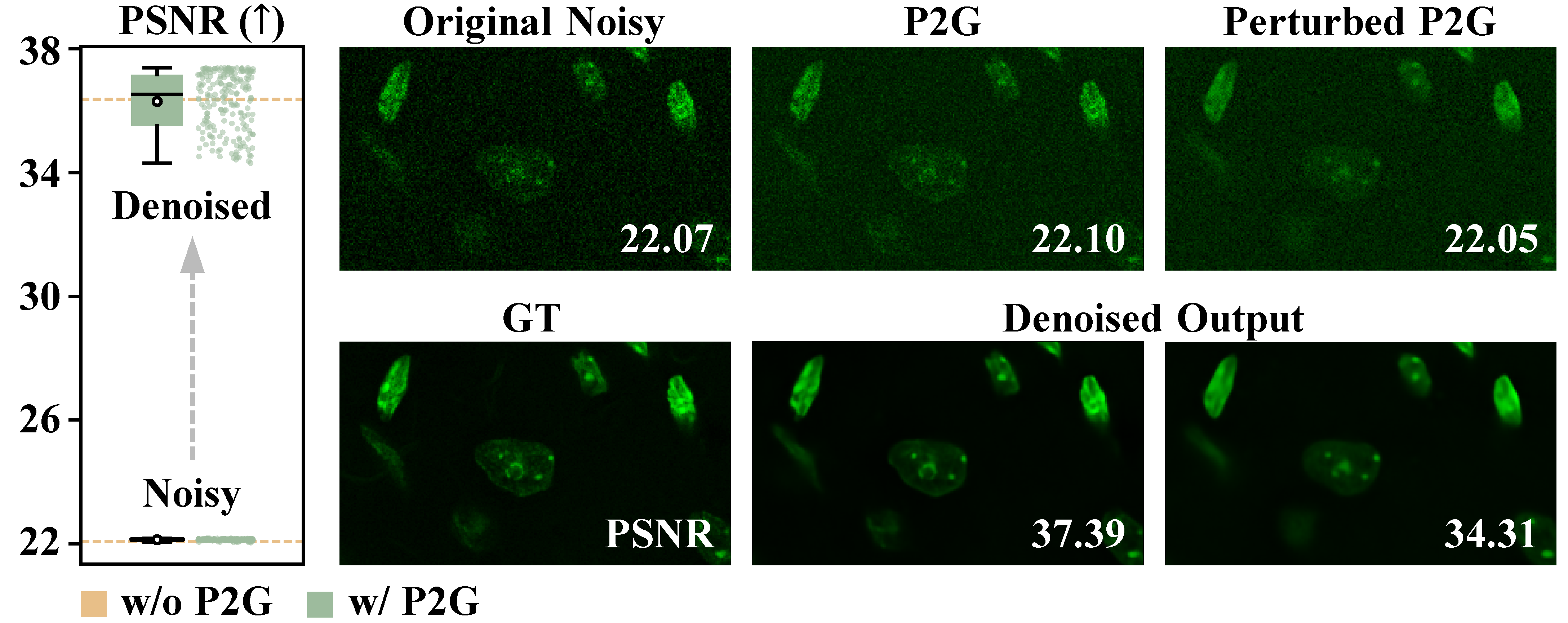}
    \caption{\textbf{Identical Gaussian likelihood does not ensure identical denoising performance.}
    Denoising PSNR under 200 post-flow Cayley orthogonal perturbations on FMDD Two-Photon Mice ($Q=16$). The learned flow and its co-adapted denoiser remain fixed; only the post-flow orthogonal mapping varies, preserving Gaussian NLL while changing denoising PSNR. The unperturbed flow ($\tau=0$), representative outputs, and ground truth are shown.}
  \label{fig5}
\end{figure}

\noindent\textbf{Comparison protocol.}
Methods are grouped by supervision: single noisy observation (1o), independent noisy pairs (2o), or external supervision/priors. P2G-1o constructs approximate pairs from one observation, while P2G-2o uses two independent noisy observations. Official SIDD results (\cref{tab:sidd}) use P2G-1o; local SIDD-Small backbone experiments (\cref{tab:backbone}) use P2G-2o. The SIDD comparison excludes methods requiring externally pretrained Gaussian denoisers, such as Noise2VST~\cite{Herbreteau2025Noise2VST}.

Results are PSNR (dB)/SSIM unless noted otherwise. Bold and underlining indicate the best and second-best results within each self-supervised group on FMDD and W2S, and among the reported methods on SIDD; ties share the corresponding mark.

N2N scores are evaluated using the pretrained checkpoint that initializes P2G-2o. On FMDD and W2S, $^*$ denotes our LAN and Noise2VST re-runs with the same initialization, splits, and noisy pairs as P2G-2o; their denoisers remain frozen during adaptation.

\noindent\textbf{Implementation.}
P2G-2o initializes its denoiser from pretrained N2N~\cite{lehtinen2018noise2noise}. For P2G-1o, we initialize from B2UB~\cite{wang2022blind2unblind} on FMDD and SIDD RAW, from B2UB and N2V~\cite{krull2019noise2void} on W2S, and from APR-RD~\cite{aprpd} on SIDD sRGB.
Approximate pairs follow N2Nbr~\cite{huang2021neighbor2neighbor}. The flow starts as the identity. Training alternates 25 flow updates ($10^{-6}$ learning rate) and 25 denoiser updates ($10^{-7}$), freezing the other component. FMDD uses $96\times96$ patches with $(\lambda_{\mathrm{mat}},\lambda_{\mathrm{NLL}})=(1,1)$; SIDD and W2S use $32\times32$ patches with loss weights $(100,1)$.

\subsection{Evaluations}\label{evaluations}

\Cref{fig4} compares a pretrained N2N denoiser with the fully adapted P2G framework across photon counts. The complete framework improves PSNR and SSIM while reducing measured marginal residual KL divergence and $1-\mathrm{SFM}$. For the representative image at $Q=16$ in \cref{fig4}b, KL decreases from 0.042 to 0.029 and $1-\mathrm{SFM}$ from 0.0029 to 0.0007. These statistics quantify marginal Gaussian agreement and spectral flatness, not full joint i.i.d. Gaussianity (Supplementary Sec.~5). The comparison measures the full framework, rather than the isolated effect of the flow.

Gaussian likelihood alone does not determine denoising utility. As shown in \cref{fig5}, 200 post-flow orthogonal perturbations preserve Gaussian NLL but change PSNR with the flow and denoiser fixed. A representative perturbation reduces PSNR from 37.39 to 34.31\,dB. With identical likelihood but different restoration outcomes (Supplementary Sec.~6), these interventions demonstrate that likelihood alone cannot select the transformation most beneficial for a fixed denoiser.

\Cref{tab:photon} provides a complementary control: transformed-input PSNR changes by only 0.04--0.06\,dB, compared with 0.34--0.96\,dB gains after full P2G training. The denoising gains substantially exceed the changes in input PSNR, highlighting the benefit of the full adaptation framework. Coupled adaptation and objective ablations in \cref{ablations} further test how the denoising objective shapes this benefit.

\begin{table}[!t]
\centering
\caption{\textbf{Input and denoised PSNR (dB) before and after P2G.} Input: original/transformed observations; output: pretrained N2N/full P2G framework.}
\label{tab:photon}
\PtwoGTableFont
\makebox[\columnwidth][c]{%
\begin{tabular*}{\PtwoGTableWidth}{@{\extracolsep{\fill}}lcccc@{}}
\otop
\textbf{} & \textbf{Q1} & \textbf{Q4} & \textbf{Q16} & \textbf{Q64} \\
\omiddle
\shortstack[l]{Noisy PSNR\\original/transformed} &
\shortstack{13.64/13.69\\\textbf{+0.05}} &
\shortstack{19.66/19.70\\\textbf{+0.04}} &
\shortstack{25.68/25.72\\\textbf{+0.04}} &
\shortstack{31.70/31.76\\\textbf{+0.06}} \\
\omiddle
\shortstack[l]{Denoised PSNR\\w/o P2G / w/ P2G} &
\shortstack{33.00/33.34\\\textbf{+0.34}} &
\shortstack{35.49/36.30\\\textbf{+0.81}} &
\shortstack{38.16/39.12\\\textbf{+0.96}} &
\shortstack{40.84/41.59\\\textbf{+0.75}} \\
\obottom
\end{tabular*}%
}
\end{table}

\subsection{Comparisons}\label{comparisons}

\noindent\textbf{Quantitative results.}
On FMDD, P2G-2o exceeds the strongest paired-noisy self-supervised baseline by 0.46--0.80\,dB across three subsets (\cref{tab:fmdd}); on W2S, it exceeds paired-noisy self-supervised baselines across all channels at avg1 and avg16, by up to 0.61\,dB at avg1 (\cref{tab:w2s}). These results demonstrate the effectiveness of the jointly adapted framework; \cref{tab:update,tab:real_cont} separate these gains from the effect of additional training.

For single-observation learning, P2G-1o performs competitively on FMDD and W2S (\cref{tab:fmdd,tab:w2s}). On SIDD, P2G-1o achieves 50.88/51.46\,dB on RAW Benchmark/Validation and 38.16\,dB on sRGB Validation, improving over B2UB~\cite{wang2022blind2unblind} by 0.09/0.10\,dB and APR-RD~\cite{aprpd} by 0.16\,dB, respectively (\cref{tab:sidd}). Over the better of each initialization and its fine-tuned counterpart, the margins are 0.06/0.08\,dB and 0.16\,dB (\cref{tab:real_cont}).

\noindent\textbf{Visual comparisons.}
\Cref{fig6} presents FMDD and SIDD RAW/sRGB examples; PSNR/SSIM are calculated on full images, rather than the displayed crops.

\begin{table}[!t]
\centering
\caption{\textbf{FMDD denoising results} (PSNR in dB / SSIM) on three microscopy subsets.}
\label{tab:fmdd}
\PtwoGTableFont
\makebox[\columnwidth][c]{%
\begin{tabular*}{\PtwoGTableWidth}{@{\extracolsep{\fill}}llccc@{}}
\otop
 & \multicolumn{1}{l}{\textbf{Method}} & \multicolumn{3}{c}{\textbf{FMDD}} \\
\cmidrule(l){3-5}
 & & \textbf{CF Fish} & \textbf{CF Mice} & \textbf{TP Mice} \\
\omiddle
Classical & BM3D+GAT \cite{dabov2007image_bm3d,makitalo2012optimal} & 32.16/0.886 & 37.93/0.963 & 33.83/0.924 \\
\omiddle
Noisy-clean & N2C \cite{Unet} & 32.79/0.905 & 38.40/0.966 & 34.02/0.925 \\
Supervision & DRUNet+GAT \cite{zhang2021dru,makitalo2012optimal} & 32.18/0.879 & 38.11/0.963 & 34.01/0.925 \\
\omiddle
Prior-assisted & Noise2VST \cite{Herbreteau2025Noise2VST} & 32.88/0.904 & 38.27/0.965 & 34.06/0.926 \\
Self-supervised & PN2V \cite{krull2020probabilistic}& 32.45/0.886 & 38.24/0.960 & 33.67/0.916 \\
\omiddle
Single-noisy & SSDN \cite{ssdn} & 31.62/0.849 & 37.82/0.959 & 33.09/0.907 \\
Self-supervised & N2F \cite{lequyer2022fast} & 32.06/\underline{0.897} & 37.73/\underline{0.965} & 33.71/\underline{0.921} \\
 & MASH \cite{Chihaoui_2024_CVPR_mash} & 30.05/0.804 & 36.55/0.958 & 32.81/0.913 \\
 & FBI-D \cite{byun2021fbi} & 32.22/0.885 & 38.32/0.963 & 33.95/0.908 \\
 & B2UB \cite{wang2022blind2unblind} & \underline{32.74}/\underline{0.897} & \underline{38.44}/0.964 & \underline{34.03}/0.916 \\
 & ZS-N2N \cite{zs-n2n} & 30.62/0.843 & 36.15/0.945 & 32.66/0.906 \\
 & N2Nbr \cite{huang2021neighbor2neighbor} & 32.11/0.890 & 37.07/0.960 & 33.40/\underline{0.921} \\
 & N2V \cite{krull2019noise2void} & 32.08/0.886 & 37.49/0.960 & 33.38/0.916 \\
 & \textbf{P2G-1o (ours)} & \textbf{32.90/0.903} & \textbf{38.51/0.967} & \textbf{34.13/0.924} \\
\omiddle
Paired-noisy & N2N \cite{lehtinen2018noise2noise} & 33.05/0.914 & 38.47/0.968 & 34.38/\underline{0.927} \\
Self-supervised & N2N (1.3$\times$ width) \cite{lehtinen2018noise2noise} & 33.06/0.914 & 38.52/0.968 & 34.20/0.926 \\
 & N2N (3.6$\times$ width) \cite{lehtinen2018noise2noise} & 32.74/0.906 & 38.02/0.966 & 33.84/0.922 \\
 & Noise2NoiseFlow \cite{noisee2noiseflow} & 32.95/0.912 & \underline{38.63}/\underline{0.969} & 33.32/0.925 \\
 & N2N+GAT \cite{lehtinen2018noise2noise,makitalo2012optimal} & 32.34/\underline{0.916} & 38.17/0.965 & 34.00/0.905 \\
 & Noise2VST* \cite{Herbreteau2025Noise2VST,lehtinen2018noise2noise} & 33.03/0.914 & 38.37/0.968 & 34.11/0.926 \\
 & LAN* \cite{kim2024lan,lehtinen2018noise2noise} & \underline{33.10}/0.915 & 38.52/0.967 & \underline{34.42}/\underline{0.927} \\
 & \textbf{P2G-2o (ours)} & \textbf{33.90}/\textbf{0.922} & \textbf{39.19/0.970} & \textbf{34.88/0.929} \\
\obottom
\end{tabular*}%
}
\end{table}

\begin{table}[!t]
\centering
\caption{\textbf{SIDD RAW Benchmark/Validation and sRGB Validation results} (PSNR in dB / SSIM). Tied best reported RAW Benchmark SSIM values are marked equally.}
\label{tab:sidd}

\PtwoGTableFont
\setlength{\arrayrulewidth}{0.7pt}

\begin{tabular*}{\columnwidth}{
@{\extracolsep{\fill}}lcc@{\hspace{4pt}}lc@{}
}

\cmidrule[0.9pt](r{2pt}){1-3}
\cmidrule[0.9pt](l{2pt}){4-5}

\textbf{Method}
& \multicolumn{2}{c}{\textbf{Raw}}
& \textbf{Method}
& \textbf{sRGB} \\

\cmidrule(lr){2-3}
\cmidrule(l){5-5}

& \textbf{Benchmark}
& \textbf{Validation}
&
& \textbf{Validation} \\

\cmidrule[0.75pt](r{2pt}){1-3}
\cmidrule[0.75pt](l{2pt}){4-5}

BM3D+GAT \cite{dabov2007image_bm3d,makitalo2012optimal}
& 48.60/0.986 & 48.92/0.986
& BM3D \cite{dabov2007image_bm3d} & 31.75/0.706 \\

N2C \cite{Unet}
& 50.60/0.991 & 51.19/0.991
& WNNM \cite{wnnm} & 26.31/0.524 \\

\cmidrule(r{2pt}){1-3}
\cmidrule(l{2pt}){4-5}

N2N \cite{lehtinen2018noise2noise}
& 50.74/\textbf{0.991} & 51.21/0.991
& CVF-SID \cite{neshatavar2022cvf} & 34.51/0.941 \\

N2V \cite{krull2019noise2void}
& 48.01/0.983 & 48.55/0.984
& AP-BSN + R$^3$ \cite{lee2022apbsn}
& 35.76/\makebox[\widthof{00.00}][c]{-} \\

ZS-N2N \cite{zs-n2n}
& 47.68/0.981 & 48.75/0.985
& LG-BPN \cite{lgbpn} & 37.32/0.886 \\

DBSN \cite{dbsn}
& 49.56/0.987 & 50.13/0.988
& SDAP \cite{Pan_2023_sdap} & 37.30/0.894 \\

SSDN \cite{ssdn}
& 50.28/0.989 & 50.89/0.990
& SASL \cite{li2023sassl} & 37.39/0.934 \\

FBI-D \cite{byun2021fbi}
& 50.57/0.990 & -
& AT-BSN (D) \cite{chen2024exploring} & 37.88/0.946 \\

B2UB \cite{wang2022blind2unblind}
& \underline{50.79}/\textbf{0.991}
& \underline{51.36}/\underline{0.992}
& APR-RD \cite{aprpd} & \underline{38.00}/\underline{0.947} \\

R2R \cite{Pang2021R2R}
& 46.70/0.978 & 47.20/0.980
& TBSN \cite{tbsn}
& 37.71/\makebox[\widthof{00.00}][c]{-} \\

N2Nbr \cite{huang2021neighbor2neighbor}
& 50.47/0.990 & 51.06/0.991
& TM-BSN \cite{Park2026TMBSN} & 37.31/0.940 \\

\textbf{P2G-1o (ours)}
& \textbf{50.88}/\textbf{0.991}
& \textbf{51.46}/\textbf{0.993}
& \textbf{P2G-1o (ours)}
& \textbf{38.16/0.953} \\

\cmidrule[0.9pt](r{2pt}){1-3}
\cmidrule[0.9pt](l{2pt}){4-5}

\end{tabular*}
\end{table}

\begin{table}[!t]
\centering
\caption{\textbf{W2S denoising results} (PSNR in dB) at avg1 / avg16 across three fluorescence channels.}
\label{tab:w2s}
\PtwoGTableFont
\makebox[\columnwidth][c]{%
\begin{tabular*}{\PtwoGTableWidth}{@{\extracolsep{\fill}}llccc@{}}
\otop
 & \multicolumn{1}{l}{\textbf{Method}} & \multicolumn{3}{c}{\textbf{W2S}} \\
\cmidrule(l){3-5}
 & & \textbf{CH0} & \textbf{CH1} & \textbf{CH2} \\
\omiddle
Classical & BM3D+GAT \cite{dabov2007image_bm3d,makitalo2012optimal} & 33.33/39.31 & 31.16/38.04 & 34.50/40.63 \\
\omiddle
Noisy-clean & DRUNet+GAT \cite{zhang2021dru,makitalo2012optimal} & 32.54/38.88 & 30.15/37.69 & 33.75/40.03 \\
Supervised & N2C \cite{Unet}& 34.30/41.94 & 32.11/39.09 & 34.73/40.88 \\
\omiddle
Prior-assisted & DivNoising \cite{DivNoising} & 34.13/39.62 & 32.28/38.37 & 35.18/40.52 \\
Self-supervised & PN2V \cite{krull2020probabilistic} & \makebox[\widthof{00.00}][c]{-} /39.19 & \makebox[\widthof{00.00}][c]{-} /38.24 & 32.48/40.49 \\
 & Noise2VST \cite{Herbreteau2025Noise2VST} & 35.65/39.74 & 33.43/39.10 & 36.88/41.36 \\
\omiddle
Single-noisy & N2V \cite{krull2019noise2void} & \underline{34.30}/\underline{38.80} & 31.80/\underline{37.81} & 34.65/\underline{40.19} \\
Self-supervised & R2R \cite{Pang2021R2R} & 28.12/38.05 & 24.97/36.74 & 25.55/37.53 \\
 & MASH \cite{Chihaoui_2024_CVPR_mash} & 30.25/35.16 & 29.47/33.21 & 32.68/37.76 \\
 & N2Nbr \cite{huang2021neighbor2neighbor} & 32.72/38.71 & 30.91/36.60 & 32.59/38.57 \\
 & CVF-SID \cite{neshatavar2022cvf} & 34.02/37.49 & 31.70/33.94 & 33.26/39.84 \\
 & ZS-N2N \cite{zs-n2n} & 32.26/37.63 & 30.56/36.59 & 32.79/39.63 \\
 & B2UB \cite{wang2022blind2unblind} & 34.22/38.32 & \underline{32.07}/35.38 & \underline{35.05}/39.73 \\
 & \textbf{P2G-1o (ours)} & \textbf{34.61}/\textbf{39.14} & \textbf{32.57}/\textbf{38.29} & \textbf{35.46}/\textbf{40.25} \\
\omiddle
Paired-noisy & N2N \cite{lehtinen2018noise2noise} & 36.52/41.38 & 34.30/39.17 & 37.00/40.77 \\
Self-supervised & N2N (1.3$\times$ width) \cite{lehtinen2018noise2noise} & \underline{37.10}/40.54 & \underline{34.87}/39.14 & \underline{37.57}/40.76 \\
 & N2N (3.6$\times$ width) \cite{lehtinen2018noise2noise} & 36.18/39.25 & 34.05/39.37 & 36.55/40.32 \\
 & Noise2NoiseFlow \cite{noisee2noiseflow} & 36.39/40.18 & 33.75/\underline{39.58} & 37.17/40.65 \\
 & N2N+GAT \cite{lehtinen2018noise2noise,makitalo2012optimal} & \makebox[\widthof{00.00}][c]{-}/38.11 & 32.47/35.99 & 35.75/39.44 \\
 & Noise2VST* \cite{Herbreteau2025Noise2VST,lehtinen2018noise2noise} & 36.54/39.59 & 32.97/38.89 & 37.00/40.23 \\
 & LAN* \cite{kim2024lan,lehtinen2018noise2noise} & 36.63/\underline{41.41} & 33.11/39.22 & 37.16/\underline{40.83} \\
 & \textbf{P2G-2o (ours)} & \textbf{37.68}/\textbf{41.71} & \textbf{35.47}/\textbf{39.66} & \textbf{38.18}/\textbf{41.16} \\
\obottom
\end{tabular*}%
}
\end{table}

\subsection{Ablations and Analysis}
\label{ablations}
We examine which components and training choices account for P2G's improvement, then test architecture transfer and computational cost. Unless otherwise noted, mechanism ablations use synthetic FMDD at $Q\in\{1,4,16,64\}$; the fine-tuning control uses the real-data benchmarks, and backbone and seed analyses use separate local splits and official SIDD RAW sets, respectively.

\noindent\textbf{Real-data fine-tuning control.}
To test whether real-data gains stem from additional training, we fine-tune each initialization with the same training pairs, matching loss, learning rate, and denoiser update count, with the flow fixed at the identity (\cref{tab:real_cont}). Fine-tuning changes PSNR by $-0.07$ to $+0.10$\,dB. P2G exceeds the better of the initialization and its fine-tuned counterpart by 0.50--0.85\,dB on FMDD, 1.16--1.18\,dB (avg1) and 0.33--0.39\,dB (avg16) on W2S, and 0.06--0.16\,dB on SIDD.

\noindent\textbf{Coupled adaptation.}
Flow-only adaptation improves PSNR by 0.20--0.80\,dB with the denoiser frozen, whereas denoiser-only updates gain at most 0.03\,dB (\cref{tab:update,tab:photon}). Coupled adaptation further improves performance, achieving gains of 0.34--0.96\,dB across photon counts.

\noindent\textbf{Denoising-oriented Gaussianization.}
We examine the roles of Gaussian NLL and denoising-oriented matching within the same two-pass framework (\cref{tab:nllweight}). With NLL alone, the flow receives no denoising feedback, and PSNR falls below pretrained N2N (\cref{tab:photon}) at $Q\le16$, by up to 1.30\,dB. Matching alone performs better but remains 0.35--0.56\,dB below the full objective despite sharing its two-pass structure. At $Q=1$, neither objective alone improves over N2N, whereas their combination gains 0.34\,dB. Together with \cref{fig5}, these results demonstrate why Gaussian likelihood must be guided by denoising feedback.

\noindent\textbf{Optimization strategy.}
Alternating updates outperform simultaneous updates by 0.18--0.44\,dB across photon counts (Supp. Tab.~S3).

\noindent\textbf{Alternative transformation methods.}
To examine whether alternative residual mappings offer comparable denoising gains, we evaluate neural, offset, and spline methods inspired by NTN~\cite{Ha2026NTN}, LAN~\cite{kim2024lan}, and Noise2VST~\cite{Herbreteau2025Noise2VST} (\cref{tab:arch}). Full P2G achieves the highest PSNR at every $Q$. Because the alternatives use frozen denoisers and different optimization budgets, and the neural variant uses a different loss, this is a method-level comparison rather than evidence that flow architecture or invertibility alone causes the gain (Supp. Sec.~7).

\noindent\textbf{Flow components.}
Removing affine coupling reduces PSNR by 0.20--0.32\,dB across photon counts, whereas the effect of CayleyMix varies with noise level (Supp. Tab.~S4).

\noindent\textbf{Backbone generality.}
To test whether the benefit depends on a particular denoiser architecture, we apply P2G-2o to UNet~\cite{Unet} and NAFNet~\cite{chen2022nafnet} with their corresponding N2N~\cite{lehtinen2018noise2noise} baselines (\cref{tab:backbone}). Across separate local SIDD-Small, FMDD, and W2S splits, P2G improves PSNR by 0.75--1.17\,dB for UNet and 0.49--0.64\,dB for NAFNet, indicating that the gains are not confined to one backbone.

\noindent\textbf{Computational cost.}
On $512\times512$ inputs (RTX 5090), P2G requires 189.6\,G FLOPs and 8.19\,ms for residual Gaussianization and two-pass denoising, compared with 45.0\,G and 3.6\,ms for N2N~\cite{lehtinen2018noise2noise}. Neither the 1.3$\times$- nor the 3.6$\times$-width N2N (77.1\,G and 580.7\,G FLOPs) matches P2G on FMDD or W2S (\cref{tab:fmdd,tab:w2s}), so the gains are not explained by model width alone. Comparisons with FBI-D, Noise2NoiseFlow, and Noise2VST are provided in Supp. Tab.~S5.

\noindent\textbf{Continuation stability.}
Across five continuation seeds sharing one P2G-1o initialization, official SIDD RAW PSNR varies by less than 0.001\,dB (Supp. Tab.~S6). This measures the stability of the adaptation stage rather than variability from independent pretraining.

\begin{table}[!t]
\centering
\caption{\textbf{Real-data fine-tuning control.}
Baseline is pretrained; Fine-tune updates only the denoiser.
FMDD/SIDD: PSNR (dB)/SSIM; W2S: PSNR (dB) at avg1/avg16.}
\label{tab:real_cont}
\PtwoGTableFont
\begin{tabular*}{\columnwidth}{@{\extracolsep{\fill}}llccc@{}}
\otop
\textbf{Dataset} & \textbf{Subset} & \textbf{Baseline} & \textbf{Fine-tune} & \textbf{P2G} \\
\omiddle
FMDD & CF Fish & \underline{33.05}/\underline{0.914} & \underline{33.05}/\underline{0.914} & \textbf{33.90}/\textbf{0.922} \\
(2o) & CF Mice & \underline{38.47}/\underline{0.968} & 38.46/\underline{0.968} & \textbf{39.19}/\textbf{0.970} \\
     & TP Mice & \underline{34.38}/\underline{0.927} & 34.37/\underline{0.927} & \textbf{34.88}/\textbf{0.929} \\
\omiddle
SIDD & RAW Benchmark & 50.79/\textbf{0.991} & \underline{50.82}/\textbf{0.991} & \textbf{50.88}/\textbf{0.991} \\
(1o) & RAW Validation   & 51.36/\underline{0.992} & \underline{51.38}/0.991 & \textbf{51.46}/\textbf{0.993} \\
     & sRGB Validation  & \underline{38.00}/\underline{0.947} & 37.93/0.946 & \textbf{38.16}/\textbf{0.953} \\
\omiddle
W2S  & CH0 & \underline{36.52}/\underline{41.38} & \underline{36.52}/41.31 & \textbf{37.68}/\textbf{41.71} \\
(2o) & CH1 & \underline{34.30}/39.17 & \underline{34.30}/\underline{39.27} & \textbf{35.47}/\textbf{39.66} \\
     & CH2 & \underline{37.00}/\underline{40.77} & \underline{37.00}/40.74 & \textbf{38.18}/\textbf{41.16} \\
\obottom
\end{tabular*}
\end{table}

\begin{table}[!t]
\centering
\caption{\textbf{Ablation of flow and denoiser adaptation.} Checkmarks indicate updated components; denoiser-only adaptation keeps the flow fixed at its identity initialization.}
\label{tab:update}
\PtwoGTableFont
\begin{tabular*}{\columnwidth}{@{\extracolsep{\fill}}cccccc@{}}
\otop
\textbf{$g_\phi$} & \textbf{$f_{\theta}$} & \textbf{Q1} & \textbf{Q4} &
\textbf{Q16} & \textbf{Q64} \\
\omiddle
$\surd$ & --
  & \underline{33.20}/0.847 & \underline{36.09}/\underline{0.901} & \underline{38.96}/\underline{0.939} & \underline{41.48}/\underline{0.960} \\
-- & $\surd$
  & 33.00/\textbf{0.860} & 35.52/\underline{0.901} & 38.18/0.937 & 40.87/0.958 \\
$\surd$ & $\surd$
  & \textbf{33.34}/\underline{0.859} & \textbf{36.30/0.908} &
    \textbf{39.12/0.942} & \textbf{41.59/0.961} \\
\obottom
\end{tabular*}
\end{table}

\begin{table}[!t]
\centering
\caption{\textbf{Ablation of flow objectives.} Checkmarks indicate the objectives used to update the flow; the denoiser is updated with matching in all settings.}
\label{tab:nllweight}
\PtwoGTableFont
\begin{tabular*}{\columnwidth}{@{\extracolsep{\fill}}cccccc@{}}
\otop
$\mathcal{L}_{\mathrm{NLL}}$ & $\mathcal{L}_{\mathrm{mat}}$ & \textbf{Q1} & \textbf{Q4} & \textbf{Q16} & \textbf{Q64} \\
\omiddle
$\surd$ & --
& 31.70/0.815 & 34.56/0.868 & 38.00/0.924 & 40.85/0.954 \\
-- & $\surd$
& \underline{32.99}/\underline{0.852} & \underline{35.85}/\underline{0.901} & \underline{38.56}/\underline{0.937} & \underline{41.07}/\underline{0.958} \\
$\surd$ & $\surd$
& \textbf{33.34/0.859} & \textbf{36.30/0.908} & \textbf{39.12/0.942} & \textbf{41.59/0.961} \\
\obottom
\end{tabular*}
\end{table}

\begin{table}[!t]
\centering
\caption{\textbf{Alternative residual-mapping methods} versus P2G (PSNR in dB / SSIM). The comparison is not budget-matched; see Supp. Sec.~7 for protocols.}
\label{tab:arch}
\PtwoGTableFont
\begin{tabular*}{\columnwidth}{@{\extracolsep{\fill}}lccccc@{}}
\otop
\textbf{Transform} & \textbf{NLL} & \textbf{Q1} & \textbf{Q4} &
\textbf{Q16} & \textbf{Q64} \\
\omiddle
NN      & --
  & 27.84/0.724 & 33.14/0.862 & 35.70/0.911 & 39.90/0.953 \\
Offset  & $\surd$
  & \underline{33.16}/\textbf{0.861} & \underline{35.59}/\underline{0.901} & \underline{38.23}/\underline{0.935} & \underline{40.88}/\underline{0.957} \\
Spline  & $\surd$
  & 32.62/0.850 & 35.18/0.892 & 37.88/0.928 & 40.62/0.955 \\
\textbf{Flow} & $\surd$
  & \textbf{33.34}/\underline{0.859} & \textbf{36.30/0.908} &
    \textbf{39.12/0.942} & \textbf{41.59/0.961} \\
\obottom
\end{tabular*}
\end{table}

\begin{table}[!t]
\centering
\caption{\textbf{Backbone generality} on local validation splits.}
\label{tab:backbone}
\PtwoGTableFont
\begin{tabular*}{\columnwidth}{@{\extracolsep{\fill}}llccc@{}}
\otop
\textbf{Architecture} & \textbf{Method} &
\textbf{SIDD} & \textbf{FMDD} & \textbf{W2S} \\
\omiddle
UNet~\cite{Unet} & N2N~\cite{lehtinen2018noise2noise}
  & \underline{45.63}/\underline{0.971} & \underline{36.79}/\underline{0.929} & \underline{35.94}/\underline{0.920} \\
 & \textbf{P2G-2o}
  & \textbf{46.45/0.984} & \textbf{37.54/0.939} & \textbf{37.11/0.937} \\
\omiddle
NAFNet~\cite{chen2022nafnet} & N2N~\cite{lehtinen2018noise2noise}
  & \underline{49.94}/\underline{0.986} & \underline{36.21}/\underline{0.913} & \underline{33.78}/\underline{0.868} \\
 & \textbf{P2G-2o}
  & \textbf{50.53/0.992} & \textbf{36.85/0.925} & \textbf{34.27/0.875} \\
\obottom
\end{tabular*}
\end{table}

\section{Discussion}
\label{sec:discussion}

P2G improves denoising and moves measured residual statistics toward i.i.d.\ Gaussian noise, but the residuals do not fully attain that distribution (\cref{fig4}). Since $Y-\hat X=(Y-X)-(\hat X-X)$, denoiser estimation errors also enter the residual transformed by P2G. Marginal KL and spectral-flatness improvements do not establish full joint i.i.d.\ Gaussianity (Supp. Sec.~5). The two-pass framework incurs additional inference cost (\cref{ablations}), while the respective contributions of noise transformation and inference structure remain to be fully disentangled.

\section{Conclusion}

Noise distributions affect the performance of in-domain self-supervised denoising even at matched noise strength. P2G turns this observation into a learnable joint residual Gaussianization transformation. Since Gaussian likelihood does not determine denoising utility, our self-supervised framework couples distribution matching with downstream denoising feedback and adapts the transformation and denoiser together. Results across controlled conditions, real datasets, and denoiser architectures support denoising-oriented Gaussianization for self-supervised restoration.

\section*{LLM Usage Statement}
We used OpenAI ChatGPT and Codex to review manuscript language and presentation and to check consistency among claims, figures, and tables. The authors remain responsible for the scientific content and final submission.

{\small
\bibliographystyle{ieeenat_fullname}
\bibliography{main}
}
\end{document}